\documentclass[11pt,a4paper]{article}
\usepackage[hyperref]{emnlp2018}
\usepackage{times}
\usepackage{latexsym}
\usepackage{amsmath}
\usepackage{graphicx}
\usepackage{caption}
\usepackage{url}

\aclfinalcopy % Uncomment this line for the final submission

\title{JUCBNMT at WMT2018 News Translation Task: Character Based Neural Machine Translation of Finnish to English}

\author{Sainik Kumar Mahata, Dipankar Das, Sivaji Bandyopadhyay\\
Computer Science and Engineering \\
	Jadavpur University, Kolkata, India \\
  {\tt sainik.mahata@gmail.com}, \\ {\tt dipankar.dipnil2005@gmail.com, sivaji\_cse\_ju@yahoo.com}}

\date{}

\begin{document}
\maketitle
\begin{abstract}
  In the current work, we present a description of the system submitted to WMT 2018 News Translation Shared task. The system was created to translate news text from Finnish to English. The system used a Character Based Neural Machine Translation model to accomplish the given task. The current paper documents the preprocessing steps, the description of the submitted system and the results produced using the same. Our system garnered a BLEU score of 12.9.
\end{abstract}

\section{Introduction}

Machine Translation (MT) is automated translation of one natural language to another using computer software. Translation is a tough task, not only for computers, but humans as well as it incorporates a thorough understanding of the syntax and semantics of both languages. For any MT system to return good translations, it needs good quality and sufficient amount of parallel corpus \cite{mahata:2016wmt2016,mahata:2017bucc2017}. 

In the modern context, MT systems can be categorized into Statistical Machine Translation (SMT) and Neural Machine Translation (NMT). SMT has had its share in making MT very popular among the masses. It includes creating statistical models, whose input parameters are derived from the analysis of bilingual text corpora, created by professional translators \cite{weaver:1955translation}. The state-of-art for SMT is Moses Toolkit\footnote{http://www.statmt.org/moses/}, created by \citet{koehn:2007moses}, incorporates subcomponents like Language Model generation, Word Alignment and Phrase Table generation. Various works have been done in SMT \cite{lopez:2008statistical,koehn:2009statistical} and it has shown good results for many language pairs. 

On the other hand NMT \cite{bahdanau:2014neural}, though relatively new, has shown considerable improvements in the translation results when compared to SMT \cite{mahataMTIL}. This includes better fluency of the output and better handling of the Out-of-Vocabulary problem. Unlike SMT, it doesn't depend on alignment and phrasal unit translations \cite{kalchbrenner:2013recurrent}. On the contrary, it uses an Encoder-Decoder approach incorporating Recurrent Neural Cells \cite{cho:2014properties}. As a result, when given sufficient amount of training data, it gives much more accurate results when compared to SMT \cite{doherty:2010eye, vaswani:2013decoding,liu:2014recursive}.

Further, NMT can be of two types, namely Word Level NMT and Character Level NMT. Word Level NMT, though very successful, suffers from a few disadvantages. It are unable to model rare words \cite{DBLP:journals/corr/LeeCH16}. Also, since it does not learn the morphological structure of a language it suffers when accommodating morphologically rich languages \cite{DBLP:journals/corr/LingTDB15}. We can address this issue, by training the models with huge parallel corpus, but, this in turn, produces very complex and resource consuming models that aren't feasible enough. 

To combat this, we plan to use Character level NMT, so that it can learn the morphological aspects of a language and construct a word, character by character, and hence tackle the rare word occurrence problem to some extent.

In the current work, we participated in the WMT 2018 News Translation Shared Task\footnote{http://www.statmt.org/wmt18/translation-task.html} that focused on translating news text, for European language pairs. The Character Based NMT system discussed in this paper was designed to accommodate Finnish to English translations. The organizers provided the required parallel corpora, consisting of 3,255,303 sentence pairs, for training the translation model. The statistics of the parallel corpus is depicted in Table \ref{Table1} Our model was trained on a Tesla K40 GPU, and the training took around 10 days to complete.
\begin{table}[h]
\centering
\begin{tabular}{|l|l|}
\hline
\textbf{\# sentences in Fi corpus} & 3,255,303 \\ \hline
\textbf{\# sentences in En corpus} & 3,255,303 \\ \hline
\textbf{\# words in Fi corpus} & 53,753,718 \\ \hline
\textbf{\# words in En corpus} & 73,694,350 \\ \hline
\textbf{\# word vocab size for Fi corpus} & 1,065,309 \\ \hline
\textbf{\# word vocab size for En corpus} & 280,822 \\ \hline
\textbf{\# chars in Fi corpus} & 427,187,612 \\ \hline
\textbf{\# chars in En corpus} & 405,624,094 \\ \hline
\textbf{\# char vocab size for Fi corpus} & 963 \\ \hline
\textbf{\# char vocab size for En corpus} & 1,360 \\ \hline
\end{tabular}
\captionsetup{justification=centering}
\caption{Statistics of the Finnish-English parallel corpus provided by the organizers. "\#" depicts No. of. "Fi" and "En" depict Finnish and English, respectively. "char" means character and "vocab" means vocabulary of unique tokens. }
\label{Table1}
\end{table}

The remainder of the paper is organized as follows. Section 2 will describe the methodology of creating the character based NMT model and will include the preprocessing steps, a brief summary of the encoder-decoder approach and the architecture of our system. This will be followed by the results and conclusion in Section 3 and 4, respectively.

\section{Methodology}
\label{method}
For designing the model we followed some standard preprocessing steps, which are discussed below.
\subsection{Preprocessing}
\label{preprocess}
The following steps were applied to preprocess and clean the data before using it for training our character based neural machine translation model. We used the NLTK toolkit\footnote{https://www.nltk.org/} for performing the steps.
\begin{itemize}
\item \textbf{Tokenization}: Given a character sequence and a defined document unit, tokenization is the task of chopping it up into pieces, called tokens. In our case, these tokens were words, punctuation marks, numbers. NLTK supports tokenization of Finnish as well as English texts.
\item \textbf{Truecasing}: This refers to the process of restoring case information to badly-cased or non-cased text \cite{lita:2003truecasing}. Truecasing helps in reducing data sparsity.
\item \textbf{Cleaning}: Long sentences (\# of tokens $>$ 80) were removed.
\end{itemize}

\subsection{Neural Machine Translation}
\label{subsec:NMT}
Neural machine translation (NMT) is an approach to machine translation that uses neural networks to predict the likelihood of a sequence of words. The main functionality of NMT is based on the sequence to sequence (seq2seq) architecture, which is described in Section \ref{subsubsec:seq2seq}.

\subsubsection{Sequence to Sequence Model}
\label{subsubsec:seq2seq}
Sequence to Sequence learning is a concept in neural networks, that helps it to learn sequences. Essentially, it takes as input a sequence of tokens (characters in our case)
\begin{equation*}
X=\{x\textsubscript{1}, x\textsubscript{2}, ..., x\textsubscript{n}\} 
\end{equation*}
and tries to generate the target sequence as output
\begin{equation*}
Y = \{y\textsubscript{1}, y\textsubscript{2}, ..., y\textsubscript{m}\}
\end{equation*}
where x\textsubscript{i} and y\textsubscript{i} are the input and target symbols respectively.

Sequence to Sequence architecture consists of two parts, an Encoder and a Decoder. 

The encoder takes a variable length sequence as input and encodes it into a fixed length vector, which is supposed to summarize its meaning and taking into account its context as well. A Long Short Term Memory (LSTM) cell was used to achieve this. The uni-directional encoder reads the characters of the Finnish texts, as a sequence from one end to the other (left to right in our case), 
\begin{equation*}
\vec{h}\textsubscript{t} = \vec{f}\textsubscript{enc}(E\textsubscript{x}(x\textsubscript{t}),\vec{h}\textsubscript{t-1})
\end{equation*}
Here, E\textsubscript{x} is the input embedding lookup table (dictionary), $\vec{f}$\textsubscript{enc} is the transfer function for the Long Short Term Memory (LSTM) recurrent unit. The cell state \textit{h} and context vector \textit{C} is constructed and is passed on to the decoder.
 
The decoder takes as input, the context vector \textit{C} and the cell state \textit{h} from the encoder, and computes the hidden state at time t as, 
\begin{equation*}
s\textsubscript{t} =  f\textsubscript{dec}(E\textsubscript{y}(y\textsubscript{t-1}), s\textsubscript{t-1}, c\textsubscript{t})
\end{equation*} 
Subsequently, a parametric function out\textsubscript{k} returns the conditional probability using the next target symbol $k$. 
\begin{equation*}
(y\textsubscript{t}=k\mid y<{t}, X) = \frac{1}{Z}exp(out\textsubscript{k}(E\textsubscript{y}(y\textsubscript{t}-1), s\textsubscript{t}, c\textsubscript{t}))
\end{equation*}
$Z$ is  the normalizing constant, 
\begin{equation*}
\sum\textsubscript{j}exp(out\textsubscript{j}(E\textsubscript{y}(y\textsubscript{t}-1), s\textsubscript{t}, c\textsubscript{t}))
\end{equation*} 
The entire model can be trained end-to-end by minimizing the log likelihood which is defined as
\begin{equation*}
L = -\frac{1}{N}\sum_{n=1}^{N}\sum_{t=1}^{T\textsubscript{y}\textsuperscript{n}}log p(y\textsubscript{t} = y\textsubscript{t}\textsuperscript{n}, y\textsubscript{<t}\textsuperscript{n}, X\textsuperscript{n})
\end{equation*}
where N is the number of sentence pairs, and X\textsuperscript{n} and y\textsubscript{t}\textsuperscript{n} are the input sentence and the t-th target symbol in the n-th pair respectively. 

The input to the decoder was one hot tensor (embeddings at character level) of English sentences while the target data was identical, but with an offset of one time-step ahead.

\subsection{Training}
\label{subsec:train}
\begin{figure}[h]
\centering
\framebox{\includegraphics[scale=0.55]{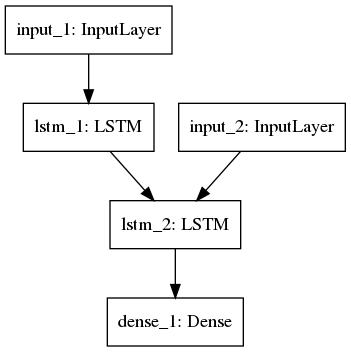}}
\caption{Architecture of the reported NMT model.}
\label{fig1}
\end{figure}

For training the model, we preprocessed the Finnish and English texts to normalize the data. Thereafter, Finnish and English characters were encoded as One-Hot vectors. The Finnish characters were considered as the input to the encoder and subsequent English characters was given as input to the decoder. A single LSTM layer was used to encode the Finnish characters. The output of the encoder was discarded and only the cell states were saved for passing on to the decoder. The cell states of the encoder and the English characters were given as input to the decoder. Lastly, a Dense layer was used to map the output of the decoder to the English characters, that were mapped with an offset of 1. The \textit{batch size} was set to 128, \textit{number of epochs} was set to 100, activation function was \textit{softmax}, optimizer chosen was \textit{rmsprop} and loss function used was \textit{categorical cross-entropy}. Learning rate was set to 0.001. The architecture of the constructed model is shown in Figure \ref{fig1}.

\section{Results}
\label{result}
Our system was a constrained system, which means that we only used data given by the organizers to train our system. The output was converted to an SGML format, the code for which was provided by the organizers. The results were submitted to \url{http://matrix.statmt.org/} for evaluation. The organizers calculated the BLEU score, BLEU-cased score, TER score, BEER 2.0 score, and Character TER score for our submission. As for the human ranking scores, the system fetched a standardized Average $Z$ score of -0.404 and a non-standardized Average $\%$ score of 58.9 \cite{bojar-EtAl:2018:WMT1}. The results of the automated and human evaluation scores are given in Table \ref{Table2}.
\begin{table}[h]
\centering
\begin{tabular}{|l|l|}
\hline
\textbf{Metrics}       & \textbf{Score} \\ \hline
\textbf{BLEU}          & 12.9           \\ \hline
\textbf{BLEU Cased}    & 12.2           \\ \hline
\textbf{TER}           & 0.816          \\ \hline
\textbf{BEER 2.0}      & 0.448          \\ \hline
\textbf{Character TER} & 0.770          \\ \hline
\textbf{Average $Z$} & -0.404          \\ \hline
\textbf{Average $\%$} & 58.9          \\ \hline
\end{tabular}
\caption{Evaluation Metrics}
\label{Table2}
\end{table}
\section{Conclusion}
The paper presents the working of the translation system submitted to WMT 2018 News Translation shared task. We have used character based encoding for our proposed NMT system. We have used a single LSTM layer as an encoder as well as a decoder. As a future prospect, we plan to use more LSTM layers in our model. We plan to create another NMT model, which takes as input words, and not characters and subsequently use various embedding schemes to improve the translation quality. 

\section*{Acknowledgments}
The reported work is supported by Media Lab Asia, MeitY, Government of India, under the Visvesvaraya PhD Scheme for Electronics \& IT.

\bibliographystyle{acl_natbib_nourl}
\bibliography{emnlp2018}
\end{document}